\pdfoutput=1

\documentclass[11pt]{article}

\usepackage[]{ACL2023}

\usepackage{times}
\usepackage{latexsym}

\usepackage[T1]{fontenc}

\usepackage[utf8]{inputenc}

\usepackage{microtype}

\usepackage{inconsolata}

\usepackage{graphicx}
\usepackage{amsmath}
\usepackage{nccmath}
\usepackage{amssymb}
\usepackage{mathtools}
\usepackage{multirow}
\usepackage{xcolor}
\usepackage{xcolor, soul}

\title{Do the Benefits of Joint Models for Relation Extraction \\
Extend to Document-level Tasks?}

\author{Pratik Saini \and Tapas Nayak \and Indrajit Bhattacharya  \\
        TCS Research, India \\
        \texttt{\{pratik.saini,nayak.tapas,b.indrajit\}}@tcs.com\\}

\begin{document}
\maketitle

\begin{abstract}
Two distinct approaches have been proposed for relational triple extraction - pipeline and joint. 
Joint models, which capture interactions across triples, are the more recent development, and have been shown to outperform pipeline models for sentence-level extraction tasks.
Document-level extraction is a more challenging setting where interactions across triples can be long-range, and individual triples can also span across sentences.
Joint models have not been applied for document-level tasks so far.
In this paper, we benchmark state-of-the-art pipeline and joint extraction models on sentence-level as well as document-level datasets.
Our experiments show that while joint models outperform pipeline models significantly for sentence-level extraction, their performance drops sharply below that of pipeline models for the document-level dataset.
\end{abstract}

\section{Introduction}

Relation extraction is a crucial NLP task for constructing and enriching knowledge bases.
Traditional pipeline approaches ~\cite{riedel2010modeling,hoffmann2011knowledge,zeng2014relation,zeng2015distant,nayak2019effective, jat2018attention} first identify entities followed by relation identification one entity pair at a time.
In contrast, more recent joint approaches~\cite{zeng2018copyre,hrlre2019takanobu,nayak2019ptrnetdecoding,Wei2020ANC,Wang2020TPLinkerSJ,Zhong2021AFE,Zheng2021PRGCPR,Li2021TDEERAE,Wei2020ANC,Yan2021APF,Shang2022OneRelJE} not only identify entities and relations for the same triple together but also extract all relational triples together. Thus, these recent approaches are better suited for capturing complex interactions.

Joint models for relation extraction outperform traditional pipeline models for sentence-level datasets such as NYT~ \cite{riedel2010modeling}.
A more natural and complex setting for relation extraction is at the document-level.
In the document-level task, relational triples may also span across sentences.
Further, there may be long range interactions between different triples across sentences.
As a result, the search space for joint models blows up with document size.
So far, research for document-level datasets such as DocRED~\cite{yao2019DocRED} has used pipeline approaches and avoided the joint approach.

In this paper, we investigate if the benefits of the joint approach extrapolate from sentence-level to document-level tasks.
We benchmark 5 SOTA joint models and 3 SOTA pipeline models on sentence-level (NYT) and document-level (DocRED) datasets.
We observe that the benefits of the SOTA joint models do not extend to document-level tasks. 
While performance of both classes of models drop sharply, joint models fall significantly below that of pipeline models. 
We perform extensive analysis to identify the short-comings of the two classes of models highlighting areas of improvement.

\begin{table*}[ht]
\small
\centering
\begin{tabular}{p{0.5in}| p{5.5in}}
\hline
Original  & \textcolor{blue}{Kiato} ( , \textcolor{blue}{Sidirodromikos Stathmos Kiatou} ) is a railway station in \textcolor{blue}{Kiato} in the northern \textcolor{red}{Peloponnese} , Greece . The station is located a kilometre west of the town , near the Greek National Road 8A ( \textcolor{red}{Patras} – Corinth highway ) . It opened on 9 July 2007 as the western terminus of the line from Athens Airport . Initially the station served as an exchange point for passengers to \textcolor{red}{Patras} on the old metre gauge SPAP line to \textcolor{red}{Patras} , but all traffic was suspended indefinitely in December 2010 for cost reasons . The nearby old \textcolor{blue}{Kiato} station was also closed . Passengers for \textcolor{red}{Patras} must now change to bus services at \textcolor{blue}{Kiato} . The station is served by one train per hour to \textcolor{red}{Piraeus} . \\ \hline
Processed & \textcolor{blue}{Kiato} ( , \textcolor{blue}{Kiato} ) is a railway station in \textcolor{blue}{Kiato} in the northern \textcolor{red}{Peloponnese} , Greece . The station is located a kilometre west of the town , near the Greek National Road 8A ( \textcolor{red}{Peloponnese} – Corinth highway ) . It opened on 9 July 2007 as the western terminus of the line from Athens Airport . Initially the station served as an exchange point for passengers to \textcolor{red}{Peloponnese} on the old metre gauge SPAP line to \textcolor{red}{Peloponnese} , but all traffic was suspended indefinitely in December 2010 for cost reasons . The nearby old \textcolor{blue}{Kiato} station was also closed . Passengers for \textcolor{red}{Peloponnese} must now change to bus services at \textcolor{blue}{Kiato} . The station is served by one train per hour to \textcolor{red}{Peloponnese} . \\ \hline
\end{tabular}
\caption{A sample document from DocRED and its processed version. The different mentions of the entity `Kiato' and `Peloponnese' are marked with \textcolor{blue}{blue} and \textcolor{red}{red} respectively. In the processed text, different mentions are normalized with the first mention of these two entities which are `Kiato' and `Peloponnese'.}
\label{tab:docred-sample-doc}
\end{table*}

\begin{table*}[ht]
\centering
\begin{tabular}{c|c|cl|cl|cl}
\hline
Dataset                & \# Relations          & \multicolumn{2}{c|}{Train}                                                                                        & \multicolumn{2}{c|}{Validation}                                                                                   & \multicolumn{2}{c}{Test}                                                                                         \\ \hline
\multicolumn{1}{l|}{} & \multicolumn{1}{l|}{} & \multicolumn{1}{l|}{\begin{tabular}[c]{@{}l@{}}\# Context \end{tabular}} & \# Triples & \multicolumn{1}{l|}{\begin{tabular}[c]{@{}l@{}}\# Context \end{tabular}} & \# Triples & \multicolumn{1}{l|}{\begin{tabular}[c]{@{}l@{}}\# Context \end{tabular}} & \# Triples \\ \hline
NYT                    & 24                    & \multicolumn{1}{c|}{56,196}                                                                           & 94,222    & \multicolumn{1}{c|}{5,000}                                                                            & 8,489     & \multicolumn{1}{c|}{5,000}                                                                            & 8,616     \\ 
DocRED                 & 96                    & \multicolumn{1}{c|}{2,572}                                                                            & 20,233    & \multicolumn{1}{c|}{284}                                                                              & 2,187     & \multicolumn{1}{c|}{924}                                                                              & 7,337     \\ \hline
\end{tabular}
\caption{Statistics of the NYT24 and DocRED datasets. Context refers to a sentence or document.}
\label{tab:data_stat}
\end{table*}

\section{Relation Extraction Approaches}



{\bf Pipeline RE approaches} solve the RE task in two sequential steps. 
In Step 1, they use an NER model to identify the entities and entity mentions in the input text.
In Step 2, they take the predicted entities and entity mentions as input, and predict all possible relations from a pre-defined relation set between pairs of entities. 
We use PL-Marker \cite{ye-etal-2022-packed} as the NER module and KD-DocRE \cite{tan-etal-2022-document}, SSAN \cite{Xu_Wang_Lyu_Zhu_Mao_2021}, and experiment with SAIS \cite{xiao-etal-2022-sais} as relation classification models for our experiments, and train these for specific datasets. 
Following standard practice, we use gold standard entity mentions for training and validation of the relation classification models, while for inference of the test instances, we naturally use the predicted entity mentions as input. 
Note that in both steps, these models perform {\em independent} classification for each entity mention and relation.


{\bf Joint RE approaches} identify the entities and relations in a relational triple in an end-to-end fashion.
Further, they consider the entire input text (sentence or document) and output a set of relational triples together, thus, capturing complex interactions across triples in theory.
The flip side, naturally, is that they need to explore a significantly larger space of candidates, which grows combinatorially with the length of the input text.
We use 5 SOTA models for experiments. 
PtrNet \cite{nayak2019ptrnetdecoding} and REBEL \cite{huguet-cabot-navigli-2021-rebel-relation} use the Seq2Seq approach. 
PtrNet generates the index position of entities in text whereas REBEL generates the tokens for the triples. 
OneRel \cite{Shang2022OneRelJE} uses a table-based tagging approach. 
The tagging approaches of BiRTE \cite{Ren2022ASB} and GRTE \cite{Ren2021ANG} have a separate entity extraction process in their end-to-end modeling. 
We train these models in an end-to-end fashion as described in the respective papers.

\section{Extraction Settings and Datasets}

The original and simpler setting for relation extraction is {\em sentence-level}. 
This setting consists of individual sentences containing one or more relations as context.
NYT \cite{riedel2010modeling} is a large-scale and popular benchmark for sentence-level RE, and we use this dataset as it is for our sentence-level experiments.


This setting, however, is restrictive since a large fraction of relations in natural text spans across multiple sentences.
This is captured in the {\em document-level} relation extraction setting. 
Here, relational triples may be intra-sentence or inter-sentence, meaning that the head and tail entities of the relational triple can span across multiple sentences and require reasoning across multiple sentences to identify them.
The task is to predict all these relations given an entire document as context.
DocRED \cite{yao2019DocRED} is a benchmark document-level dataset that we use for our experiments.

Contexts in DocRED (avg. number of tokens around 197) are much longer than in NYT (avg. number of tokens around 37). 
However, training data size is much larger for NYT.
Since the relation labels of the DocRED test set are not released, we use the original validation set as test set and split the training data for training and validation. 
DocRED has mostly been used for pipeline models.
We needed additional processing to make it tractable for joint models. 
We remove the documents with overlapping entity mentions in the training and validation set. 
We get 2,856 documents from the training set and 924 documents from the validation set. 
Then, we replace all entity mentions with the first occurring entity mention for each entity so that co-reference resolution is not required. We include an example of document processing for DocRED in Table \ref{tab:docred-sample-doc}. 
The details of the dataset splits of the NYT and DocRED for our experiments are included in Table \ref{tab:data_stat}\footnote{Our processed DocRED dataset is available at https://github.com/pratiksaini4/nyt-docred-joint-pipeline-comparison}. 

\textbf{Evaluation Metric:} We use `strict' criteria for evaluation. We consider an extracted relational triple as correct only if two entities and relation exactly match with a ground truth triple. We report triple level precision, recall and F1 scores for the models.

\textbf{Parameter Settings:} We use $BERT_{BASE}$ (cased) \cite{Devlin2019BERTPO} for document encoding for all the models except REBEL for which we have used $BART_{BASE}$ \cite{Lewis2019BARTDS}. We used NVIDIA Tesla V100 32GB GPU to train the models. We used other hyper-parameters as per provided in their respective papers.




\section{Results and Discussion}

\begin{table*}[ht]
\centering
\begin{tabular}{c|c|lll|lll}
\hline
\multicolumn{1}{l|}{}            & \multicolumn{1}{l|}{} & \multicolumn{3}{c|}{NYT24}                                              & \multicolumn{3}{c}{DocRED}                                            \\ \cline{2-8} 
\multicolumn{1}{l|}{}            & Model                 & \multicolumn{1}{c}{P} & \multicolumn{1}{c}{R} & \multicolumn{1}{c|}{F1} & \multicolumn{1}{c}{P} & \multicolumn{1}{c}{R} & \multicolumn{1}{c}{F1} \\ \hline
\multirow{5}{*}{Joint Models}    & OneRel                & 0.926                 & 0.918                 & 0.922                   & 0.513                 & 0.130                 & 0.208                  \\
                                 & BiRTE                 & 0.914                 & 0.920                 & 0.917                   & 0.522                 & 0.402                 & 0.454                  \\
                                 & GRTE                  & 0.929                 & 0.924                 & 0.926                   & 0.586                 & 0.373                 & 0.456                  \\
                                 & PtrNet                & 0.898                 & 0.894                 & 0.896                   & 0.222                 & 0.145                 & 0.175                  \\
                                 & Rebel                 & 0.881                 &  0.885                &   0.883                   & 0.466                 & 0.356                 & 0.404                  \\ \hline
\multirow{3}{*}{Pipeline Models} & KD-DocRE              & 0.895                 & 0.910                 & 0.902                   & 0.620                 & 0.556                 & 0.586                  \\
                                 & SSAN                  & 0.781                 & 0.798                 & 0.789                   & 0.576                 & 0.529                 & 0.552                  \\
                                 & SAIS                  & 0.864                 & 0.879                 & 0.872                   & 0.640                 & 0.545                 & 0.589                  \\ \hline
\end{tabular}
\caption{Performance of the SOTA models on end-to-end relation extraction on NYT24 and DocRED datasets.}
\label{tab:sota-performance}
\end{table*}

\begin{table*}[ht]
\centering
\begin{tabular}{c|l|lll|lll}
\hline
\multicolumn{1}{l|}{} &
  \multicolumn{1}{c|}{\multirow{2}{*}{Model}} &
  \multicolumn{3}{c|}{Intra} &
  \multicolumn{3}{c}{Inter} \\ \cline{3-8} 
\multicolumn{1}{l|}{} &
  \multicolumn{1}{c|}{} &
  \multicolumn{1}{c}{P} &
  \multicolumn{1}{c}{R} &
  \multicolumn{1}{c|}{F1} &
  \multicolumn{1}{c}{P} &
  \multicolumn{1}{c}{R} &
  \multicolumn{1}{c}{F1} \\ \hline
\multirow{2}{*}{Joint Models}    & BiRTE    & 0.600 & 0.425 & 0.497 & 0.420 & 0.366 & 0.391 \\
                                 & GRTE     & 0.677 & 0.407 & 0.508 & 0.460 & 0.320 & 0.378 \\ \hline
\multirow{2}{*}{Pipeline Models} & KD-DocRE & 0.666 & 0.601 & 0.631 & 0.545 & 0.485 & 0.513 \\
                                 & SAIS     & 0.697 & 0.594 & 0.641 & 0.548 & 0.467 & 0.504 \\ \hline
\end{tabular}
\caption{Performance of top performing SOTA models on Intra vs Inter relational triples on DocRED dataset.}
\label{tab:intra-inter-performance}
\end{table*}

Through our experiment, we try to find out the answers to following research questions (RQ).


\noindent{\bf RQ1: How do the two classes of models perform at sentence and documents scales?}

End-to-end performance of the joint models and the pipeline models on the NYT and DocRED datasets is shown in Table~\ref{tab:sota-performance}. 
On the sentence-level NYT dataset, both joint models and pipeline models achieve close to 0.90 F1 score. 
But, on DocRED, we see a huge drop in the F1 score for both categories.
The pipeline models score below 0.60 whereas among the joint models GRTE and BiRTE perform around 0.45, the others drop to 0.20 or below.
One reason for the drop for DocRED is the smaller training data size.
But, it does not explain the gap of 10\% F1 score between the pipeline and joint models when they performed almost at par for NYT.
This suggests that joint models struggle with longer context and cross-sentence relations of documents.
We investigate this in more detail next.

\noindent{{\bf RQ2: Are some joint models better than others at document scale?}

Out of the 5 joint models, we see significantly higher drop in F1 score for OneRel and PtrNet than REBEL, GRTE, and BiRTE models. Given a document with L tokens and K predefined relations, OneRel maintains a three-dimensional matrix $M_{L \times K \times L}$ and assigns tags for all possible triples. 
When context length grows as in documents, M has many more negative tags and very few positive tags, which seems to affect OneRel performance. 
BiRTE and GRTE, on the other hand, extract the entities first separately and then classify the relations. While this is done in an end-to-end fashion, it is still similar to pipeline approach. 
This may be the reason for the smaller performance drop compared to pipelines models on DocRED.

PtrNet and REBEL are Seq2Seq models which use a decoder to extract the triples, so they possibly need more training data to learn from longer document contexts. 
Additionally, PtrNet extracts index positions for the entities. 
Since an entity may appear more than once in a document, we mark the first occurring index of the entity-mention to train this model. This very likely contributes to its poorer performance. 
On the other hand, REBEL outputs entities as text, and does not have this training issue.

\noindent{{\bf RQ3: How different are performances for intra vs inter-sentence extraction?}

The fundamental difference between NYT and DocRED is that NYT contains only intra-sentence triples, whereas DocRED contains both intra and inter-sentence (cross-sentence) triples. 
In Table \ref{tab:inter-intra-dist}, we first show intra vs inter sentence relational triple distribution for the gold and model predictions on the DocRED dataset. 
Pipeline models have nearly the same distribution for gold and prediction. 
But, joint models are skewed towards intra-sentence relations. 
This suggests that joint models are very good at extracting intra-sentence triples but they struggle with inter-sentence triples. 
This is why joint models perform very well on the NYT dataset and fail to do so on DocRED.

\begin{table}[ht]
\small
\centering
\begin{tabular}{l|l|cc}
\hline
\multirow{2}{*}{} & \multicolumn{1}{c|}{\multirow{2}{*}{Model}} & \multicolumn{2}{c}{Predict}                                 \\ \cline{3-4} 
                  & \multicolumn{1}{c|}{}                       & \multicolumn{1}{l}{Intra \%} & \multicolumn{1}{l}{Inter \%} \\ \hline
\multirow{1}{*}{-}    & Gold    & 61 & 39 \\ \hline
\multirow{2}{*}{Joint Models}    & BiRTE    & 56 & 44 \\
                                 & GRTE     & 58 & 42 \\ \hline
\multirow{2}{*}{Pipeline Models} & KD-DocRE & 62 & 38 \\
                                 & SAIS     & 62 & 38 \\ \hline
\end{tabular}
\caption{Intra vs Inter relational triple distribution of SOTA models predictions. We include top two models from the joint and pipeline class for this analysis.}
\label{tab:inter-intra-dist}
\end{table}

\begin{table}[ht]
\small
\centering
\begin{tabular}{c|cc|cc}
\hline
\multirow{2}{*}{\# Hops} & \multicolumn{2}{c|}{Pipeline Models} & \multicolumn{2}{c}{Joint Models} \\ \cline{2-5} 
  & KD-DocRE & SAIS & BiRTE & GRTE \\ \hline
1 & 0.50     & 0.45 & 0.38  & 0.33 \\
2 & 0.44     & 0.44 & 0.34  & 0.31 \\
3 & 0.48     & 0.49 & 0.36  & 0.31 \\
4 & 0.48     & 0.49 & 0.37  & 0.31 \\
5 & 0.51     & 0.54 & 0.39  & 0.32 \\
6 & 0.42     & 0.43 & 0.29  & 0.32 \\ \hline
\end{tabular}
\caption{Recall score of Pipeline and Joint SOTA models on inter-sentence relations with respect to distance between subject and object entities.}
\label{tab:hop-wise-recall}
\end{table}

In Table \ref{tab:intra-inter-performance}, we have reported the intra-sentence vs inter-sentence relations performance of the top-performing models on the DocRED test dataset. We see that all the models perform way better at intra-sentence extraction as compared to inter-sentence extraction, it demonstrates that inter-sentence extraction is significantly harder. 
We also see that pipeline models achieve around 10\% higher F1 scores than the joint models for {\em both} intra and inter categories on DocRED. 
This shows that even for the familiar intra-sentence setting joint models face more difficulties compared to pipeline models when encountered with longer context and smaller training volume.

Lastly, we investigate the impact of the distance between subject and object mentions in the context on the performance of inter-sentence relations.
In Table \ref{tab:hop-wise-recall}, we record recall of SOTA models on inter-sentence relations for different subject-object hop distances. 
Hop distance $k$ refers to the minimum sentence-level distance between the subject and object entity of a triple within the document being $k$.
Again, we see that pipeline models outperform joint models by $\sim12\%$ for all hop distances, and not just for longer ones. 




\noindent{\bf RQ4: How is performance affected by training data size?}

Next, we analyze how training volume affects performance for the two model classes for the simpler intra-sentence extraction task.
Note that NYT contains such relations exclusively.
Since DocRED has both categories, we prepare DocRED-Intra including only intra-sentence triples and the corresponding sentences.
The size of these datasets are significantly different. 
DocRED-Intra has only $\sim6.5K$ training instances compared to $94K$ for NYT.
We train all the models with these intra-sentence triples and record their performance for DocRED-Intra in Table \ref{tab:sent-intra-performance}. 
Corresponding NYT performance is in Table \ref{tab:sota-performance}.
We observe a big gap of $\sim42- 50\%$ for joint models and $\sim33\%$ for pipeline models in the performance between NYT and DocRED-Intra. 
This is due to the smaller training volume associated with a larger number of relations in DocRED.
The notable disparity between pipeline and joint models in the case of DocRED-Intra demonstrates that joint models are not as effective at generalization compared to pipeline models, particularly when working with limited training data volumes and longer contexts.


\begin{table}[ht]
\small
\centering
\begin{tabular}{c|c|ccc}
\hline
                                 
                                 &  Model                      & P       & R       & F1          \\ \hline
\multirow{2}{*}{Joint Models} & BiRTE & 0.527 & 0.462 & 0.492  \\
                              & GRTE  & 0.544 & 0.346 & 0.423  \\ \hline
\multirow{2}{*}{Pipeline Models} & KD-DocRE               & 0.524   & 0.619   & 0.567     \\
                                 & SAIS                   & 0.485   & 0.610   & 0.540     \\ \hline
\end{tabular}
\caption{Performance of top performing SOTA models on DocRED-Intra dataset.}
\label{tab:sent-intra-performance}
\end{table}


\noindent{\bf RQ5: How different are entity extraction performances at sentence and document scales?}

Finally, we aim to analyze if the huge gap in the performance of pipeline and joint models on DocRED is affected by their performance on NER subtask of relation extraction.
In Table \ref{tab:ner-performance}, we include the performance of these models on the entity extraction task. 
Pipeline models have a separate NER model. The performance of this model - PL-Marker - on NER task is similar to that of the BiRTE and GRTE models for NYT dataset. 
But, for DocRED, BiRTE and GRTE perform much worse than PL-Marker, the drop in F1 score being around 25\%. 
This, in turn, hurts their performance on the relational triple extraction. 
Though training data volume for DocRED is smaller, note that the PL-Marker model is trained on the DocRED dataset itself as are the BiRTE/GRTE models. 
This shows that, aside from overall extraction performance, joint models struggle with the NER subtask as well when training data is limited.
This suggests that a separate NER model may be more useful in such settings.

\begin{table}[ht]
\small
\centering
\begin{tabular}{c|c|lll}
\hline
Dataset                  & Model     & \multicolumn{1}{c}{P} & \multicolumn{1}{c}{R} & \multicolumn{1}{c}{F1} \\ \hline
\multirow{3}{*}{NYT24}   & PL-Marker & 0.948                 & 0.955                 & 0.952                  \\
                         & BiRTE     & 0.955                 & 0.954                 & 0.954                  \\
                         & GRTE      & 0.958                 & 0.956                 & 0.957                  \\ \hline
\multirow{3}{*}{DocRED} & PL-Marker & 0.942                 & 0.934                 & 0.938                  \\
                         & BiRTE     & 0.73                  & 0.647                 & 0.686                  \\
                         & GRTE      & 0.757                 & 0.576                 & 0.654                  \\ \hline
\end{tabular}
\caption{Performance of PL-Marker, BiRTE and GRTE on the NER task for NYT24 and DocRED dataset.}
\label{tab:ner-performance}
\end{table}




\section{Conclusion}

While joint models for relational triple extraction have been shown to outperform pipeline models for sentence-level extraction settings, in this paper we have demonstrated, with extensive experimentation, that these benefits do not extend to the more realistic and natural document-level extraction setting, which entails longer contexts and cross-sentence relations.
Experimenting with 5 SOTA joint models and 3 SOTA pipeline models, we have shown that while performance of both classes of models drops significantly in the more complex document setting, joint models suffer significantly more with longer context, inter-sentence relations and limited training data for the overall task as well as for subtasks such as NER.
This aids in establishing a research agenda for joint models to extend the promised benefits of joint entity identification, relation classification, and joint extraction of all triples from the context. This pertains to the more challenging yet natural and crucial setting for relation extraction.

\section{Limitations}

The major limitation of this work is that we could only analyze 3 pipeline models and 5 joint models. Recently, many models have been proposed for this task both in pipeline and joint class. Out of these, we chose different kinds of SOTA models to cover the different design choices made by these models. We chose PtrNet \cite{nayak2019ptrnetdecoding} and REBEL \cite{huguet-cabot-navigli-2021-rebel-relation} as they used Seq2Seq model for this task. OneRel \cite{Shang2022OneRelJE} used table-filling method whereas BiRTE \cite{Ren2022ASB} and GRTE \cite{Ren2021ANG} used sequentially extracting entities and relations in their end-to-end model.

\bibliography{anthology,custom}
\bibliographystyle{acl_natbib}

\appendix

\section{Appendix}
\label{sec:appendix}

\subsection{Details of Joint Models}

\subsubsection{PtrNet \cite{nayak2019ptrnetdecoding}}
PtrNet utilizes a seq2seq approach along with the pointer network-based decoding for jointly extracting entities and relations. Each triple contains the start and end index of the subject and object entities, along with the relation class label. Their decoder has two-pointer networks to identify the start and end index of the two entities and a classifier to identify the relation between the two entities. Decoder extracts a relational triple at each time steps and continue the process till there is no triple to extract. To ensure parity with other SOTA models, their BiLSTM encoder is replaced with BERT encoder.

\subsubsection{REBEL \cite{huguet-cabot-navigli-2021-rebel-relation}}
REBEL utilizes an auto-regressive seq2seq model that streamlines the process of relation extraction by presenting triples as a sequence of text and uses special separator tokens, as markers, to achieve the linearization.. WordDecoder model of \cite{nayak2019ptrnetdecoding} uses a similar approach using LSTMs whereas REBEL is a BART-based Seq2Seq model that utilizes the advantages of transformer model and pre-training. REBEL uses a more compact representation for the relational triples over WordDecoder model.

\subsubsection{GRTE \cite{Ren2021ANG}}
GRTE utilizes individual tables for each relation. The cell entries of a table denote the presence or absence of relation between the associated token pairs. It uses enhanced table-filling methods by introducing two kinds of global features. The first global feature is for the association of entity pairs and the second is for relations. Firstly, a table feature is generated for each relation, which is then consolidated with the features of all relations. This integration produces two global features related to the subject and object, respectively. These two global features are refined multiple times. Finally, the filled tables are utilized to extract all relevant triples.

\subsubsection{OneRel \cite{Shang2022OneRelJE}}
OneRel frames the joint entity and relation extraction task as a fine-grained triple classification problem. It uses a three-dimensional matrix with relation-specific horns tagging strategy. The rows in this matrix refer to the head entity tokens and the columns in this matrix refer to the tail entity tokens from the original text. The scoring-based classifier checks the accuracy of the decoded relational triples. It discards the triples with low confidence.

\subsubsection{BiRTE \cite{Ren2022ASB}}
In this paper, a bidirectional tagging approach with multiple stages is utilized. BiRTE first discovers the subject entities and then identifies the object entities based on the subject entities. It then does this in the reverse direction, first discovering the object entities and identifying subject entities for the object entities. The final stage involves the relation classification of subject-object pairs. They perform these tasks jointly in a single model.

\subsection{Details of Pipeline Models}

\subsubsection{SSAN \cite{Xu_Wang_Lyu_Zhu_Mao_2021}}
This paper frames the structure of entities as defined by the specific dependencies between the entity mention pairs in a document. The proposed approach, SSAN, integrates these structural dependencies with the self-attention mechanism at the encoding stage. To achieve this, two transformation modules are included in each self-attention building block, which generates attentive biases to regulate the attention flow adaptively. This approach achieved SOTA performance on the DocRED dataset.

\subsubsection{KD-DocRE \cite{tan-etal-2022-document}}
This paper suggests a semi-supervised framework for extracting document-level relations. To achieve this, they exploit the inter-dependency among the relational triples through the implementation of an axial attention module. This approach leads to enhanced performance when dealing with two-hop relations. In addition to this, an adaptive focal loss is proposed as a means of resolving the issue of imbalanced label distribution for long-tail classes. Finally, to account for the difference between human-annotated data and distantly supervised data, knowledge distillation is utilized. Their experiments on the DocRED show the effectiveness of the approach.

\subsubsection{SAIS \cite{xiao-etal-2022-sais}}
The objective of this paper is to train the model to identify relevant contexts and entity types by using the Supervising and Augmenting Intermediate Steps (SAIS) approach for relation extraction. The SAIS framework proposed in this paper results in the extraction of relations that are of superior quality, owing to its more efficient supervision. By utilizing evidence-based data augmentation and ensemble inference, SAIS also improves the accuracy of the supporting evidence retrieval process while minimizing the computational cost.

\subsection{Details of NER Model}

\subsubsection{PL-Marker \cite{ye-etal-2022-packed}}
In their approach for span representation, PL-Marker strategically uses levitated markers to consider the interrelation between pairs of spans. They propose a packing strategy that factors in neighbouring spans to improve the modeling of entity boundary information. Additionally, they utilize a subject-oriented packing approach, which groups each subject with its objects to effectively model the interrelations between the same subject span pairs.

\subsection{Related Work}


\textbf{Sentence-level Relation Extraction:} Early approaches for relation extraction use two steps pipeline approach. The first step, Named Entity Recognition (NER), extracts entities from the text. The second step, Relation Classification (RC), identifies pairwise relations between the extracted entities \cite{zeng2014relation,zeng2015distant,jat2018attention,nayak2019effective}. Pipeline methods fail to capture the implicit correlation between the two sub-tasks. They suffer from error propagation between the two stages. They cannot model the interaction among the relational triples.


To mitigate the drawbacks of pipeline approaches, recent works have focused on Joint entities and relation extraction. Joint approaches referred to as End-to-End Relation Extraction (RE) accomplish both tasks jointly. Training simultaneously on both NER and RC tasks allows for capturing more complex interactions among the multiple relational triples present in the context. \citet{miwa2016end} proposed a model that trained the NER and RC module in a single model. \citet{nayak2019ptrnetdecoding, Cabot2021REBELRE} propose seq2seq models for extracting the triples in a sequence. \citet{DBLP:journals/corr/abs-2011-01675} casts the joint extraction task as a set prediction problem rather than a sequence extraction problem. \citet{zhang-etal-2017-end, Wang2020TPLinkerSJ, wang-etal-2021-unire, Shang2022OneRelJE} formulate the NER and RC tasks as table filling problem where each cell of the table represents the interaction between two tokens.   \citet{Ren2022ASB, Zheng2021PRGCPR, Li2021TDEERAE, Yan2021APF, Wei2020ANC} have separate NER and RC modules in the same model trained in an end-to-end fashion.

\textbf{Document-level Relation Extraction:} Recently, there has been a shift of interest towards document-level RE \cite{yao2019DocRED}. Document-level relation extraction (DocRE) is known to be a more complex and realistic task compared to the sentence-level counterpart. DocRE typically involves large volumes of data, which can be computationally expensive to handle using joint models. Recent work in DocRE has avoided using joint models for this task as joint models are not scalable for long documents.
In DocRE, there can be multiple mentions of an entity with different surface forms across the document and the evidence of the relations can spread across multiple sentences. In document-level RE, mostly pipeline approaches are proposed, joint extraction approaches are not explored for this task. Earlier works \cite{peng-etal-2021-cross, quirk-poon-2017-distant, song-etal-2018-n, jia-etal-2019-document} used dependency graph between the two entities to find the relations. Recent works \cite{guo-etal-2019-attention, nan-etal-2020-reasoning, wang-etal-2020-global, zeng-etal-2020-double, zeng-etal-2021-sire, Xu_Chen_Zhao_2021, xu-etal-2021-discriminative} proposed graph-based approaches that use advanced neural techniques to do multi-hop reasoning. More recent Transformer-based approaches \cite{DBLP:journals/corr/abs-1909-11898, DBLP:journals/corr/abs-2003-12754, huang-etal-2021-entity, Xu_Wang_Lyu_Zhu_Mao_2021, Zhou_Huang_Ma_Huang_2021, xie-etal-2022-eider} use pre-trained language models to encode long-range contextual dependencies in the documents. \citet{huang-etal-2021-entity, xie-etal-2022-eider, xiao-etal-2022-sais, tan-etal-2022-document} use neural classifier to identify the evidences for relations along with relation classification for performance improvement. 

\end{document}